\documentclass[runningheads]{llncs}

\usepackage{eccv}

\usepackage{eccvabbrv}

\usepackage{graphicx}
\definecolor{bestcolor}{HTML}{f69d9d}
\definecolor{sbestcolor}{HTML}{ffeab6}
\definecolor{tbestcolor}{HTML}{fdffba}
\newcommand{\best}[1]{\cellcolor{bestcolor} #1}
\newcommand{\sbest}[1]{\cellcolor{sbestcolor} #1}
\newcommand{\tbest}[1]{\cellcolor{tbestcolor} #1}
\usepackage{multirow}
\usepackage{booktabs}
\usepackage{makecell}

\usepackage[accsupp]{axessibility}  % Improves PDF readability for those with disabilities.

\usepackage[pagebackref,breaklinks,colorlinks,citecolor=eccvblue]{hyperref}
\usepackage{orcidlink}

\begin{document}

% ---------------------------------------------------------------
% TODO REVIEW: Replace with your title
\title{Human Mesh Recovery from  Arbitrary Multi-view Images} 

% TODO REVIEW: If the paper title is too long for the running head, you can set
% an abbreviated paper title here. If not, comment out.
% \titlerunning{Abbreviated paper title}

% TODO FINAL: Replace with your author list. 
% Include the authors' OCRID for the camera-ready version, if at all possible.
\author{Xiaoben Li\inst{1,2}\and
Mancheng Meng\inst{2} \and
Ziyan Wu\inst{2} \and Terrence Chen\inst{2} \and Fan Yang\inst{2}\thanks{Corresponding author.} \and Dinggang Shen\inst{1,2}}

% TODO FINAL: Replace with an abbreviated list of authors.
\authorrunning{X.~Li et al.}
% First names are abbreviated in the running head.
% If there are more than two authors, 'et al.' is used.

% TODO FINAL: Replace with your institution list.
% \institute{ShanghaiTech University \and
% Springer Heidelberg, Tiergartenstr.~17, 69121 Heidelberg, Germany
% \email{lncs@springer.com}\\
% \url{http://www.springer.com/gp/computer-science/lncs} \and
% ABC Institute, Rupert-Karls-University Heidelberg, Heidelberg, Germany\\
% \email{\{abc,lncs\}@uni-heidelberg.de}}

\institute{ShanghaiTech University
\and United Imaging Intelligence \\
\email{\{lixb1, mengmch, dgshen\}@shanghaitech.edu.cn}\\
\email{\{ziyan.wu, terrence.chen, fan.yang03\}@uii-ai.com}
}
\maketitle

\begin{figure}
    \centering
    \includegraphics[width=\linewidth]{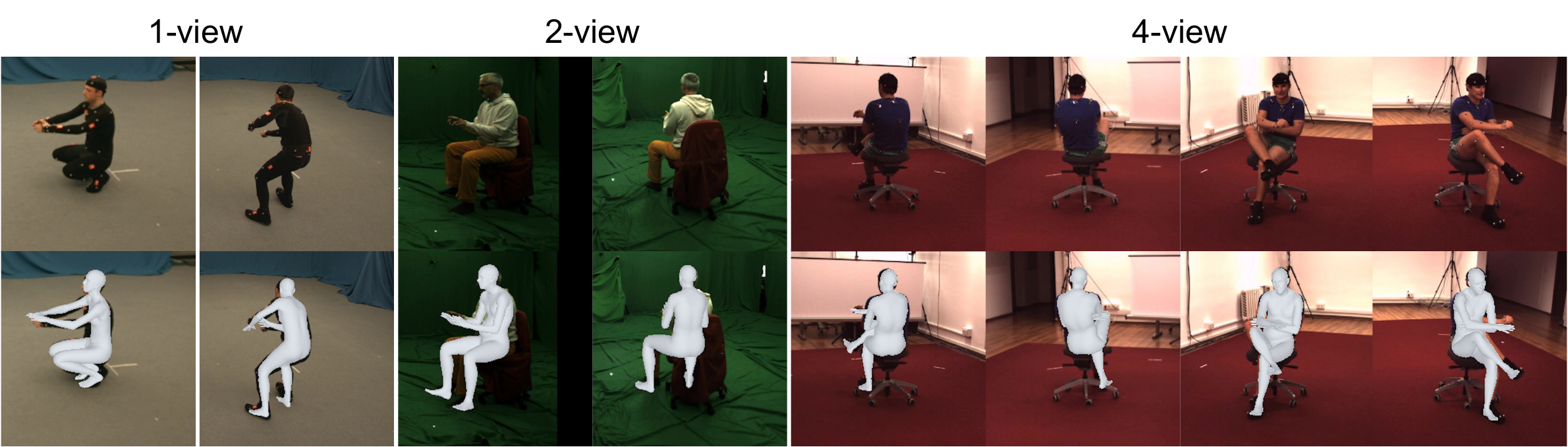}
    % \caption{{\bf Human mesh recovery from arbitrary multi-view images.} We propose a concise, flexible, and effective framework for human mesh recovery from arbitrary multi-view images. From left to right the results of mesh recovery on 1-view, 2-view, and 4-view images (from three different datasets) are shown respectively. Our framework can be directly adapted to arbitrary number of views without any modification, fine-tuning or re-training, and can learn multi-view information effectively for human mesh recovery. For the limitation of page space, up to 4-view results are displayed here. Results on more views are in supplementary.}
    \caption{{\bf Unified Human Mesh Recovery (U-HMR): Recovering human pose and shape from arbitrary multi-view images.} We propose a concise, flexible, and effective framework for human mesh recovery from arbitrary multi-view images. From left to right the results of human mesh recovery from 1-view, 2-view, and 4-view images (from three different datasets) are shown respectively. Our framework can be directly adapted to arbitrary number of views without any modification, fine-tuning or re-training, and can learn multi-view information effectively for human mesh recovery. For the limitation of page space, up to 4-view results are displayed here. Results on more views are in supplementary material.}
    \label{fig:teaser}
\end{figure}
\vspace{-1em}
\begin{abstract}
  Human mesh recovery from arbitrary multi-view images involves two characteristics: the arbitrary camera poses and arbitrary number of camera views. Because of the variability, designing a unified framework to tackle this task is challenging. The challenges can be summarized as the dilemma of being able to simultaneously estimate arbitrary camera poses and recover human mesh from arbitrary multi-view images while maintaining flexibility. 
  % Specifically, the estimation of camera poses and cross-view fusion for mesh recovery 
  % \fnote {may need to swap the position of ``camera pose" and ``cross-view fusion" to make it align with following text} are typically dependent on the number of views, making it difficult to design a single framework for arbitrary multi-view images. To detach this dependency, 
  To solve this dilemma, we propose a divide and conquer framework for Unified Human Mesh Recovery (U-HMR) from arbitrary multi-view images. In particular, U-HMR consists of a decoupled structure and two main components: camera and body decoupling (CBD), camera pose estimation (CPE), and arbitrary view fusion (AVF). As camera poses and human body mesh are independent of each other, CBD splits the estimation of them into two sub-tasks for two individual sub-networks (\ie, CPE and AVF) to handle respectively, thus the two sub-tasks are disentangled. In CPE, since each camera pose is unrelated to the others, we adopt a shared MLP to process all views in a parallel way. In AVF, in order to fuse multi-view information and make the fusion operation independent of the number of views, we introduce a transformer decoder with a SMPL parameters query token to extract cross-view features for mesh recovery. To demonstrate the efficacy and flexibility of the proposed framework and effect of each component, we conduct extensive experiments on three public datasets: Human3.6M, MPI-INF-3DHP, and TotalCapture.  Code is available at: \url{https://github.com/XiaobenLi00/U-HMR}.
\end{abstract}

\section{Introduction}
\label{sec:intro}
% \begin{figure} 
%     \centering
%     \includegraphics[width=\linewidth]{figures/teaser.pdf}
%     \caption{We split the task of reconstructing 3D human pose and shape from multi-view images into two parts: 1) the estimation of body pose and shape parameters, 2) the estimation of camera parameters. We decouple these two parts and estimate the two sets of parameters separately. The camera parameters of each view are estimated independently from the corresponding image features, and the body pose and shape parameters are estimated after fusing the information from all views.}
%     \label{fig:teaser}
% \end{figure}

\begin{figure*} 
    \centering
    \includegraphics[width=0.98\linewidth]{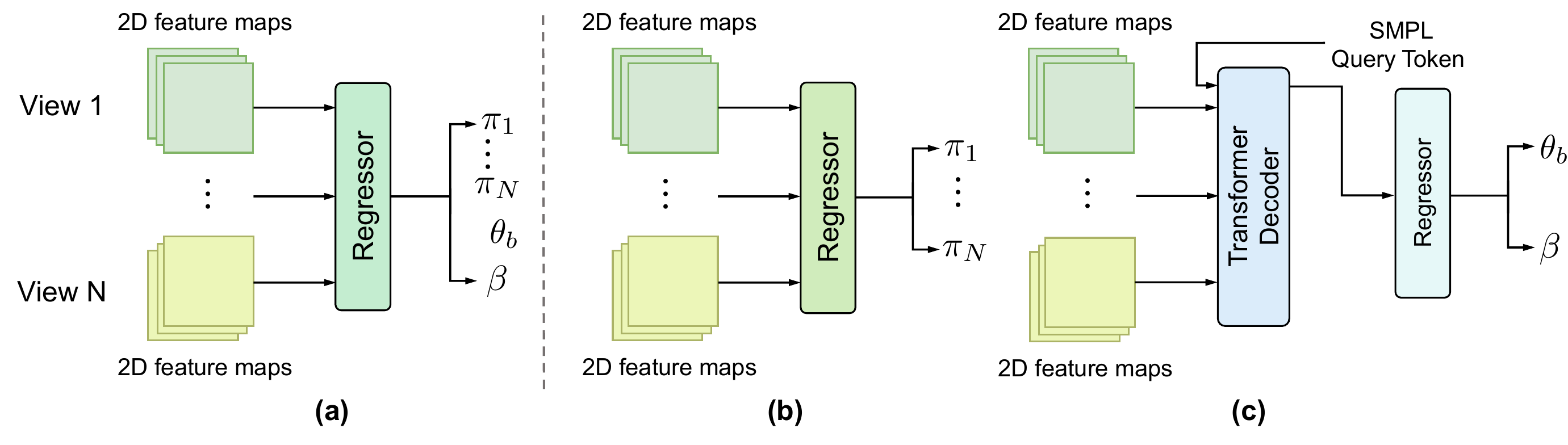}
    \caption{{\bf The comparison of multi-view scenario with a coupled structure (left) and arbitrary multi-view  scenario with a decoupled structure (right).} (a): The conventional model structure for human mesh recovery from multi-view image. (b): The model structure of camera pose $\pi_i$ estimation for arbitrary number of views, note that the regressor here is shared across different views. (c): The model structure of arbitrary multi-view feature fusion for body pose $\theta_b$ and shape $\beta$ estimation.
    % \fnote{In abstract we present AVF first, then CPE. But here we introduce CPE first.}
%     % \fnote{the figure (b) and (c) looks too similar with figure 3. And this figure is a little large.}
    }
    \label{fig:idea}
\end{figure*}

\vspace{-1em}
Human mesh recovery~\cite{bogo2016smplify, kanazawa2018hmr, pavlakos2018learning} as a fundamental task in computer vision involves estimating both the pose and shape of the human body and producing mesh representation of the human body from input visual data. It has a wide range of applications, including human-computer interaction, augmented reality, and virtual reality. Due to the critical importance and broad applications, numerous works have been conducted in this field.

Considerable amount of efforts are spent on recovering human mesh from a single image~\cite{bogo2016smplify, kanazawa2018hmr, pavlakos2018learning, kolotouros2019spin, kocabas2021pare, goel20234dhumans}. Although those methods can recovery human mesh from any single-view image, it is an ill-posed task because of depth ambiguity and possible occlusion. As multi-view images inherently contain more comprehensive information and multi-camera systems that can capture multi-view images are becoming more common, it is more feasible for human mesh recovery. 

Considering that multi-view images contain crucial information for human mesh recovery, such as camera poses, human pose and shape from each view, and cross-view 3D geometry, many works~\cite{liang2019mv-shapeaware, shin2020mv-lva,yu2022mv-uncalibrated, jia2023mv-paff} have been conducted to explore the usability of the above information. However, most of them are restricted to scenarios with fixed camera poses or number of views. The structure of these methods can be summarized into \cref{fig:idea} (a).
% While some methods can be applied to arbitrary multi-view scenario, their model structures are complicated in order to effectively leverage the aforementioned multi-view information. This may hinder flexibility when scaling up to different number of views.
While some methods can also be applied to arbitrary multi-view scenarios~\cite{yu2022mv-uncalibrated,jia2023mv-paff}, they focus on multi-view fusion and their model structures are complicatedly designed in order to effectively leverage the aforementioned multi-view information but not paying much attention to the flexibility.
% This may hinder flexibility when scaling up to different number of views.
More importantly, to the best of our knowledge, no attempts have been made to systematically investigate how to design a concise and unified network architecture to efficiently and effectively accommodate arbitrarily varied camera poses and number of views for human mesh recovery.

We follow the HMR~\cite{kanazawa2018hmr, goel20234dhumans} direction to recover human mesh. Designing a concise framework for arbitrary multi-view images is a challenging task, because the “arbitrary” contains two properties: 1) the poses of cameras are arbitrary, 2) the number of views is arbitrary. The arbitrary poses of cameras lead to the requirement of simultaneous camera pose estimation and mesh recovery, which increases the difficulty of model learning. Moreover, since both camera pose estimation and human mesh recovery from multi-view images are closely related to the number of views, this further rises the difficulty of dealing with arbitrary multi-view images with a single network.  

We observe that the camera poses and human body mesh (pose and shape) are naturally uncorrelated to each other. In addition, each camera pose is independent of others, so that estimating one camera pose is actually identical to estimating arbitrary number of camera poses. Therefore, if we can separate camera pose estimation and human mesh recovery (\cref{fig:idea} right), at least the camera pose estimation task can be relatively easy to achieve independence from number of views (\cref{fig:idea} (b)).

In terms of the multi-view feature fusion for human mesh recovery, the flexibility of fusion network structure is constrained by the number of views. Inspired by the query based transformer decoder structure~\cite{carion2020detr,liu2022petr,goel20234dhumans}, the multi-view fusion can be modeled by using a query token to integrate feature from different locations and views
% into a fused feature
for body pose and shape estimation (\cref{fig:idea} (c)). In this way, the structure of fusion network is not influenced by the number of views.  

Based on the analysis above, we propose a divide and conquer strategy to deal with the difficulties. Specifically, we construct a concise network architecture, named Unified Human Mesh Recovery network (U-HMR). The network is mainly composed by a camera and body decoupling structure (CBD), a camera pose estimation module (CPE), and an arbitrary view fusion module (AVF). CBD divides the human mesh recovery into two sub-tasks: camera pose estimation and body pose/shape estimation, for two independent sub-networks to handle. Consequently, it can mitigate the model learning difficulty. In addition, 3D human pose and shape is independent of the camera pose, CBD is intuitively aligned with this observation. As the camera poses of different views are not correlated with each other, we adopt a weight-shared MLP in CPE to process camera poses in a parallel manner. To alleviate the dependency on number of views in multi-view fusion, we introduce a transformer decoder with a SMPL parameters query token in AVF, so that the cross-view image features can be aggregated regardless of the varying number of views. Furthermore, it also work well in a single view scenario because of its flexibility for arbitrary number of views.

Our contributions can be summarized as follows:
\begin{enumerate}
\item 
We comprehensively investigate how to design a concise and unified framework for human mesh recovery from arbitrary multi-view images, and propose a divide and conquer architecture by decoupling camera pose estimation and human mesh recovery.
\item 
We propose to introduce transformer decoder with a SMPL parameters query token to aggregate the multi-view feature.
\item We conduct extensive experiments on large datasets to validate the flexibility and efficacy of our method and proposed components.

    % \item We split the task of reconstructing 3D human pose and shape into two parts: 1) the estimation of body pose and shape parameters, 2) the estimation of camera parameters. We decouple these two parts and estimate the two sets of parameters separately.
    % \item We use a simple transformer decoder-based framework to fuse information from multiple views without the need of camera calibrations. The framework is flexible and can be easily extended to incorporate arbitrary number of views.
    % \item We conduct extensive experiments on large datasets to validate the efficacy and flexibility of our method.
\end{enumerate}
\section{Related Work}
\label{sec:relatedwork}
% \subsection{HMR}
\noindent
{\bf Human Mesh Recovery for a Single Image.}
Works in this field are roughly divided into two categories: optimization-based methods and regression-based methods. Optimization-based methods~\cite{sigal2007combined,guan2009estimating,bogo2016smplify,lassner2017unite,zanfir2018monocular,pavlakos2019expressive, guler2019holopose,xiang2019monoculartotalcapture,tiwari22posendf}
usually fit a parametric human body model, \eg, SMPL~\cite{loper2015smpl}, to input image by iteratively minimizing the difference between the projected 3D mesh and the 2D observations,
% These approaches attempt to advance this field from various directions: different human body model, \eg, SCAPE~\cite{anguelov2005scape} and SMPL~\cite{loper2015smpl}, different 2D cues including 2D keypoints and silhouettes~\cite{lassner2017unite,zanfir2018monocular,xiang2019monoculartotalcapture,guler2019holopose}, and different prior used for optimization~\cite{pavlakos2019expressive,tiwari22posendf}.
% Early work~\cite{sigal2007combined,guan2009estimating} attempts to fit the SCAPE model~\cite{anguelov2005scape} to 2D silhouettes or keypoints. SMPLify~\cite{bogo2016smplify} is the first fully automatic optimization-based method that fits the SMPL model~\cite{loper2015smpl} to 2D keypoints leveraging multiple strong priors. Beyond SMPLify, methods~\cite{lassner2017unite,zanfir2018monocular,xiang2019monoculartotalcapture,guler2019holopose} involve multiple cues including 2D keypoints and silhouettes for optimization. Pavlakos \etal~\cite{pavlakos2019expressive} propose SMPLify-X which use a new neural network-based pose prior and a new interpenetration penalty to fit SMPL-X model that extends SMPL with fully articulated hands and an expressive face. Rencently, Tiwari \etal~\cite{tiwari22posendf} propose Pose-NDF which also be used as a prior in optimization-based 3D pose estimation from images. 
% Despite the many optimization-based methods, regression-based methods still dominate.
while regression-based methods~\cite{kanazawa2018hmr,pavlakos2018learning,kolotouros2019graphcmr,moon2020i2l-meshnet,kocabas2021spec,kocabas2021pare,zhang2021pymaf,li2022cliff,zheng2023potter,zheng2023feater,ma2023virtualmarkers} use deep neural networks to directly regress the 3D mesh from input image. In the scope of this paper we focus on regression-based methods. Starting from HMR~\cite{kanazawa2018hmr}, numerous methods have been proposed to improve the performance of human mesh recovery. Most of them investigate more comprehensive feature representation for the regression of SMPL parameters. However, the performance of the methods are still constrained due to the inherent depth ambiguity and possible occlusion of single image.

% In addition to these two categories of methods, many methods address this problem from other perspectives, \eg, non-parametric representation of human body~\cite{kolotouros2019graphcmr,lin2021metro,lin2021meshgraphormer,cho2022fastmetro,dou2023tore}, probabilistic distribution modeling of human body~\cite{kolotouros2021prohmr,sengupta2021probabilistic,sengupta2021hierarchical,sengupta2023humaniflow,fang2023propose}, the usage of inverse kinematics~\cite{georgakis2020hkmr,iqbal2021kama,li2021hybrik,yu2021skeleton2mesh,li2023niki,shetty2023pliks}, the collaboration of optimization-based and regression-based methods~\cite{kolotouros2019spin, li2021mv-spin,joo2021eft,moon2022neuralannot} and so on.

As transformer~\cite{vaswani2017transformer} has been successfully used in many computer vision applications~\cite{dosovitskiy2020vit,carion2020detr,he2021mae}, it is also used for human mesh recovery. METRO~\cite{lin2021metro} and Graphormer~\cite{lin2021meshgraphormer} introduce transformer encoder to model global interactions among mesh vertices and joints. Following METRO, several methods are proposed to reduce computational cost~\cite{cho2022fastmetro,dou2023tore,zheng2023potter,zheng2023feater} or leverage pixel-aligned features~\cite{yoshiyasu2023deformer,kim2023sampling}. While METRO and its followers mostly focus on non-parametric mesh recovery, transformers can also be used in parametric mesh recovery. Yang~\etal~\cite{yang2023capturing} propose a novel Transformer-based model with a design of independent tokens which are updated to estimate SMPL parameters conditioned on a given image. OSX~\cite{lin2023osx} proposes one-stage framework with component-aware transformer to capture the connections of body parts and output whole-body model parameters~\cite{pavlakos2019expressive}. In 4DHumans~\cite{goel20234dhumans}, HMR2.0, a fully ``transformerized'' version of network for human mesh recovery, is proposed. HMR2.0 uses ViT~\cite{dosovitskiy2020vit} as the image encoder and a standard transformer decoder with multi-head self-attention to produce human model parameters. Our method benefits from advance of transformer as well by using a transformer decoder-based framework to fuse multi-view information.
For a more comprehensive review, we refer readers to a survey on recovering 3D human mesh from monocular images~\cite{tian2023hmrsurvey}.
%~\cite{kanazawa2018hmr,kolotouros2019spin,kocabas2021spec,kocabas2021pare,kolotouros2021prohmr,zhang2023pymafx,loper2015smpl}

\noindent
{\bf Human Mesh Recovery from Multi-view Images.}
% Although multi-view images provide more information compared to single view image, reconstructing 3D human pose and shape from multi-view images is nontrivial, as information from multi-view images should be complementary and fused appropriately. 
% Compared to single-view image,
Multi-view images inherently contain more comprehensive information, making it more feasible for human mesh recovery~\cite{liang2019mv-shapeaware, shin2020mv-lva,yu2022mv-uncalibrated, chun2023mv-lmt,jia2023mv-paff}. Since more works focuse on multi-view 3D human pose estimation (3DHPE) and can inspire multi-view human mesh recovery, we include some works on this topic here. When camera calibrations (intrinsics and extrinsics) are available, multi-view information can be integrated using multi-view geometry. In~\cite{iskakov2019learnable,shin2020mv-lva,chun2023mv-lmt}, 2D features from multi-view images are back-projected and aggregated into a 3D feature volume using camera calibrations. 3D geometry information can also be fused into 2D features~\cite{he2020epipolartransformers,ma2021transfusion} using epipolar geometry for multi-view 3D human pose estimation. But camera calibration is not easy to obtain, especially in some dynamic scenes. When camera calibrations are unavailable, multi-view information is usually fused via learning based methods. In \cite{shuai2022adaptive, zhou2023efficient}, multi-view information is fused by Transformer for multi-view 3D human pose estimation without camera calibrations. As for human mesh recovery, Liang~\etal~\cite{liang2019mv-shapeaware} use a recurrent module to predict human model and camera parameters view by view and stage by stage. Yu~\etal~\cite{yu2022mv-uncalibrated} leverage the human body as a semantic calibration target, in which way multi-view features are aligned and fused without the need of camera calibrations. Instead of focusing on multi-view information fusion, our work is to solve the entanglement of body mesh, camera pose, and arbitrary multi-view in estimating SMPL parameters.
% Instead of designing domain-specific methods, we propose a decoupling structure and a transformer decoder-based network to fuse information from arbitrary multiple views for human mesh recovery without the need of camera calibrations.
% \label{sec:method}

% \externaldocument{2_relatedwork}
\section{Method}
\label{sec:method}
\begin{figure*}[t]
    \centering
    \includegraphics[width=0.98\linewidth]{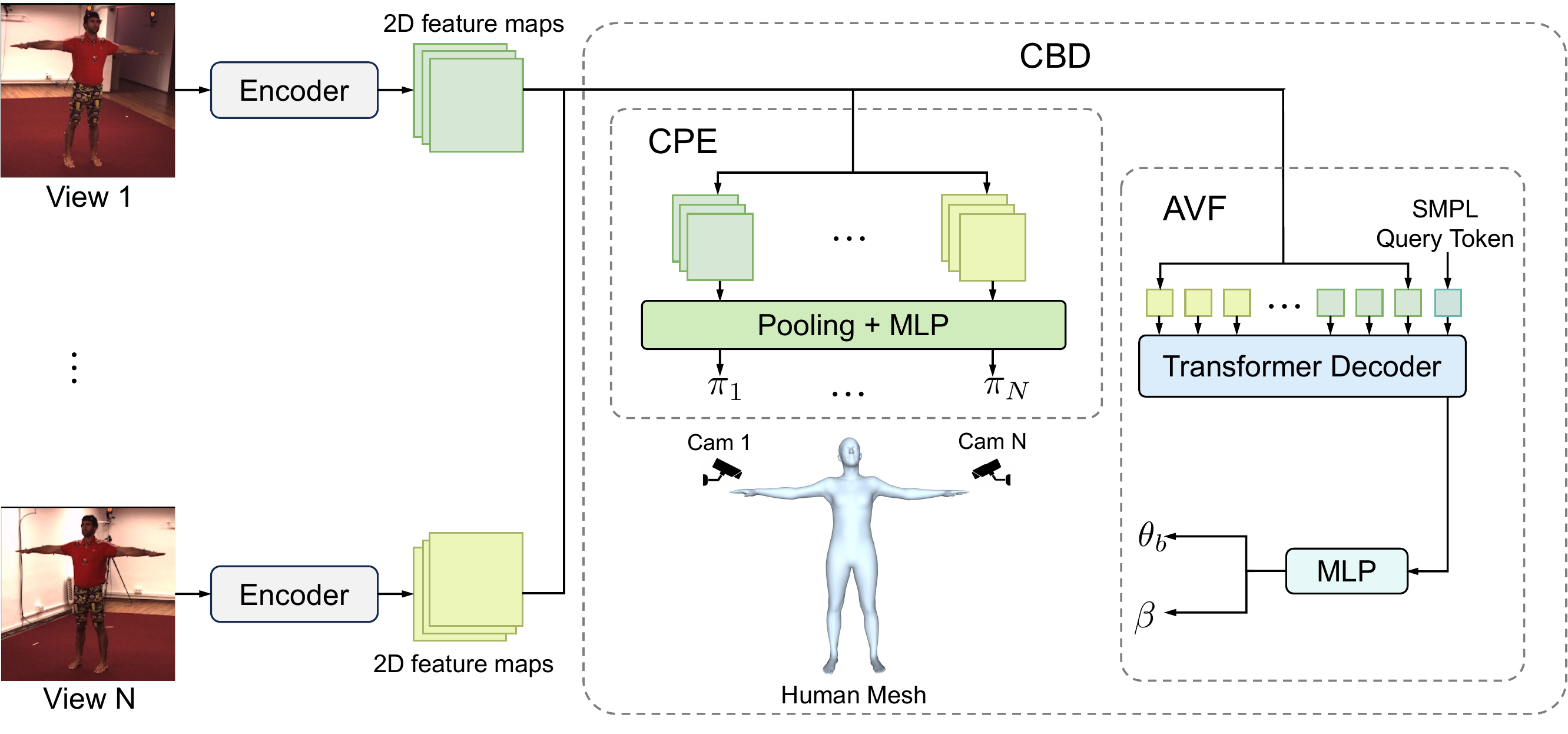}
    \caption{{\bf Overview of the proposed framework.} We divide the task of reconstructing 3D human mesh from arbitrary multi-view images into two sub-tasks: 1) the estimation of camera parameters, 2) the estimation of body mesh (pose~\&~shape) parameters. This is achieved by a  camera and body decoupling structure (CBD).
    % We decouple these two sub-tasks and estimate the two sets of parameters separately.
    Given $N$ images of a human from different camera views, we first extract 2D features from each image using a 2D image encoder. Then the 2D feature maps are forwarded to two modules, camera pose estimation module (CPE), and arbitrary view fusion module (AVF). In CPE, feature maps of each view are fed into an MLP, which is shared across views, to predict camera parameters $\pi_i$ of each view independently. In AVF, feature maps from all views are reshaped into tokens and forwarded into a transformer decoder. Inspired by PETR~\cite{liu2022petr}, multi-view position embeddings are adopted to distinguish the tokens of different regions and views. 
    A single learnable SMPL query token is introduced to attend tokens from multi-view images to form a cross-attention structure, so that the multi-view information is implicitly encoded into the SMPL query token
    % explicitly weighted summed into a feature vector 
    which is subsequently used to produce final body pose parameters $\theta_b$ and shape parameters $\beta$. 
    % \fnote{M ay need to modify the figure a little bit so that it is consistent with the text in method section. For instance, CBD, CPE, and AVF are not directly noted in the figure.}
    % \fnote{may need to lift "the body and pose" a little bit so that it is aligned with "view N", this can decrease the blank margin of the figure.}
    }
    % \vspace{-1.5em}
    \label{fig:framework}
\end{figure*}

\subsection{Human Body Representation}
% \noindent
% {\bf Body Model.}
The SMPL (Skinned Multi-Person Linear Model) model~\cite{loper2015smpl} is a skinned vertex-based model and can accurately represent a wide variety of body shapes 
.
% in natural human poses.
It provides a function $M(\theta, \beta;\Phi): \mathbb{R}^{|\theta|\times|\beta|}\mapsto \mathbb{R}^{3N}$, mapping pose and shape parameters to $N=6,890$ vertices to form a mesh representation of human body. In the function $\theta \in \mathbb{R}^{24\times 3}$, $\beta \in \mathbb{R}^{10}$, and $\Phi$ are pose, shape, and model parameters, respectively. The pose parameters $\theta$ includes global rotation $\theta_g \in \mathbb{R}^{3}$ which is unrelated to the body pose and body pose parameters $\theta_b \in \mathbb{R}^{23\times 3}$, consequently $\theta = \{\theta_g, \theta_b\}$. The shape parameters $\beta$ is coefficient of a low-dimensional shape space, learned from a training set of thousands of registered scans.

% \vspace{-2em}
Typically, a pre-trained linear regressor $W$ is used to obtain $k$ keypoints 
% of interest 
of the whole body, \ie, $J_{3D}\in \mathbb{R}^{k\times 3} = WM(\theta, \beta;\Phi)$~\cite{kolotouros2019spin}. As for 2D keypoints, a perspective camera model with fixed focal length and intrinsics $K$ is used to project 3D joint positions into 2D image plane. Each camera pose $\pi = \{R, t\}$ consists of a global orientation $R \in \mathbb{R}^{3}$ and a translation $t\in\mathbb{R}^3$.
Given above parameters, 3D keypoints can be projected to 2D image as  $J_{2D} = \Pi (K(RJ_{3D}+t))$, where $\Pi$ is a perspective projection with camera intrinsics $K$. Since the pose parameters $\theta$ already includes a global orientation $\theta_g$ which can represent the camera rotation, in practice we consider $R=\theta_g$, so when reconstructing human body from a single image, we estimate body pose parameters $\theta_b$ and camera pose  $\pi = \{R, t\}$, and when reconstructing human body from $N$ images, we estimate body pose parameters $\theta_b$ and $N$ camera poses  ${\pi_i = \{R_i, t_i\}}_{i=1}^N$.
% $R$ of the camera model as identity matrix and only predict camera translation $t$.
% \noindent
% {\bf Camera Model.}
\subsection{Unified Human Mesh Recovery (U-HMR)}
\noindent
{\bf Overview.}
The overall architecture of proposed  Unified Human Mesh Recovery network (U-HMR) is depicted in \cref{fig:framework}. Generally, U-HMR is composed by
% 1. an arbitrary view feature extraction module, 
1. a 2D image encoder,
2. a camera and body decoupling structure (CBD), 3. a camera pose estimation module (CPE), and 4. an arbitrary view fusion module (AVF). Given images of a human body from $N$ camera views, we first extract 2D features $f_i$ from each image $I_i$ using a shared 2D image encoder $E$, where $f_i=E(I_i)$. The instantiation of the encoder can be any visual fundamental model (\eg, ResNet, ViT). These features are simultaneously forwarded into CPE and AVF modules via a bifurcating operation in CBD. CPE predicts all camera poses (${\pi_i,...,\pi_N}$) with a view-invariant MLP network. AVF integrates cross-view features with a transformer decoder by introducing a SMPL query token which attends tokens from multi-view images and is forwarded
% into a fused feature and feeds the feature 
into an MLP to output human body pose $\theta_b$ and shape $\beta$. Note that even in single view scenario, the proposed framework can work well.

\noindent
{\bf Camera and Body Decoupling (CBD).}
The 3D human body pose and shape are independent of camera poses. Therefore, it is natural and intuitively reasonable to divide the camera pose estimation and human mesh recovery to two individual sub-tasks. Moreover, in arbitrary multi-view scenario, both camera pose estimation and human mesh recovery are heavily related with the number of camera views. This entanglement results in the rising difficulty of dealing with arbitrary number of views. By decoupling the two tasks, we can handle the arbitrary problem in a divide-and-conquer manner and make it easier to eliminate the dependence of sub-tasks on the number of camera views. 

\noindent
{\bf Camera Pose Estimation (CPE).} While the computation complexity of estimating camera pose is linearly correlated with the number of views, the camera pose of each view is independent of others. Consequently, the model complexity can be detached from the number of views. Specifically, we adopt a cross-view shared MLP network to predict all camera poses in a parallel way. In this way, the network structure of estimating camera pose is completely disentangled with the number of views, so as to adapt to arbitrary multi-view scenario. The formally expression of the whole CPE module is:
\begin{equation}
    \begin{aligned}
    \{\pi_i\}_{i=1}^N&= \{F_c(f_i)\}_{i=1}^N
\end{aligned}
\end{equation}
where $f_i$, $\pi_i$, and $F_c$ denote the image feature, camera pose, and MLP, and $N$ is the number of views.

\noindent
{\bf Arbitrary View Fusion (AVF).}
In terms of fusing arbitrary multi-view features for human mesh recovery, we expect the following two properties: 1) the complementary information from different views should be effectively fused,
% integrated into a single feature vector;
2) the fusion operation should be unrelated with the number of views. To achieve these properties, we introduce a transformer decoder-based architecture with a query token to aggregate cross-view features. 
Specifically, given 2D features $ \{f_i\}_{i=1}^N$ from all views, we first reshape them into tokens and feed them into a transformer decoder $\Omega$. Inspired by PETR~\cite{liu2022petr}, multi-view position embeddings are added to the tokens to distinguish tokens from different regions and views. Furthermore, to fuse the information of different regions and views, we introduce a single learnable SMPL query token, $q$, to attend tokens from multi-view images.
After the cross-attention layers in the transformer decoder, the updated SMPL query token $\hat{q}$ is expected to encode the information from tokens of multi-view images.
% and form a fused feature.
Then a light-weight MLP, $F_b$, predicts the pose parameters $\theta_b$ and shape parameters $\beta$ from  the updated SMPL query token $\hat{q}$. The computation process from the 2D features to the predicted parameters can be expressed as follows:
\begin{equation}
\label{eq:encoder}
    \begin{aligned}
    &\hat{q} = \Omega (q, \{f_i + p_i\}_{i=1}^N)\\
    &\{\theta_b, \beta\} = F_b(\hat{q})
\end{aligned}
\end{equation}
Essentially, the multi-view feature fusion is a process of weighted sum of each view feature. In the transformer decoder structure, because of the cross-attention mechanism, the number of the weights is independent of the number of the input views. Due to this online variability, we can fuse features from arbitrary multi-view images. In addition, benefiting from the feature representation learning ability of the transformer decoder, the  the updated SMPL query token can be more representative for human mesh recovery.       

\subsection{Loss Functions}
% \noindent
% {\bf Losses.} 
Following the best practice of the HMR~\cite{kanazawa2018hmr} and HMR2.0~\cite{goel20234dhumans}, we use a combination of 2D reprojection losses, 3D losses including 3D keypoints losses and SMPL parameters losses, and adversarial losses provided by a discriminator. During the training process, they are empirically weighted. The loss functions are slightly different in multi-view setting. When the ground-truth SMPL pose parameters $\hat{\theta}$ and shape parameters $\hat{\beta}$ are available, we supervise the model predictions using a squared L2 loss:
% \fnote{in my understanding, L2 is squared. }
% \xnote{L2-norm itself is squared off: $||X||_2=\sqrt{\sum_{i=1}^N|x_i^2|}$}
% \fnote{not sure this is a mean squared error, looks like a L2 loss.}
% \xnote{MSE loss is the means squared error averaged by the number of samples, squared L2-norm is identical to MSE loss for a sample and often used with `reduction = mean', it is indeed a squared L2 loss so we can change the name.}
    % $$\mathcal{L}_{smpl} = ||\theta - \hat{\theta}||_2^2 + ||\beta - \hat{\beta}||_2^2.$$
    \begin{equation*}
        \mathcal{L}_{smpl} = ||\theta - \hat{\theta}||_2^2 + ||\beta - \hat{\beta}||_2^2.
    \end{equation*}
When accurate ground-truth 3D keypoints $\hat{J}_{3D}$ are available in the dataset, a L1 loss is adopted to measure the difference of predicted 3D keypoints $J_{3D}$ with $\hat{J}_{3D}$. Since camera parameters for each view are estimated, we can use them to transform 3D keypoints from SMPL coordinate system to the coordinate system of each camera, \ie, our predicted 3D keypoints $J_{3D}$ are in the camera coordinate system. The total loss is the sum of losses from all views:
% $$\mathcal{L}_{3D} = \sum_i^N||J_{3D}^i - \hat{J}_{3D}^i||_1.$$
\begin{equation*}
    \mathcal{L}_{3D} = \sum_i^N||J_{3D}^i - \hat{J}_{3D}^i||_1.
\end{equation*}
When the dataset provides accurate 2D keypoints annotations $\hat{J}_{2D}$, we supervise the model training with a L1 loss of the 2D projection of predicted 3D keypoints $J_{2D}$ with respect to $\hat{J}_{2D}$. 2D reprojection loss is often considered to encourage pixel-level alignment. Similar to the 3D loss, the total 2D loss is the sum of losses from all views:
% $$\mathcal{L}_{2D} = \sum_i^N||J_{2D}^i - \hat{J}_{2D}^i||_1.$$
\begin{equation*}
    \mathcal{L}_{2D} = \sum_i^N||J_{2D}^i - \hat{J}_{2D}^i||_1.
\end{equation*}
In addition, we want to ensure that our model predicts valid 3D poses and use the adversarial prior proposed in HMR~\cite{kanazawa2018hmr} to regularize the prediction. It factorizes the model parameters into: (i) body pose parameters
% $\theta$,
$\theta_b$,
(ii) shape parameters $\beta$, and (iii) per-part relative rotations $\theta_i$, which is one 3D rotation for each of the 23 joints of the SMPL model. A discriminator $D_k$ is trained for each factor of the body model, where $k$ denotes the index of factors. The generator loss can be expressed as following:
% \fnote{did we really use this loss?}
% $$\mathcal{L}_{adv} = \sum_k(D_k(\theta_b, \beta) - 1)^2.$$
\begin{equation*}
    \mathcal{L}_{adv} = \sum_k(D_k(\theta_b, \beta) - 1)^2.
        % \mathcal{L}_{adv} = \sum_k(D_k(\theta, \beta) - 1)^2.
\end{equation*}
% \noindent
% {\bf Architecture.}
\section{Experiments}
\label{sec:exps}
\subsection{Experimental Setup}
\noindent
{\bf Datasets.}
We evaluate our method on three large multi-view motion capture datasets, \ie,  Human3.6M~\cite{ionescu2014h36m}, MPI-INF-3DHP~\cite{mehta2017mpi-inf-3dhp}, and TotalCapture~\cite{trumble2017totalcapture}. Particularly, on Human3.6M we comprehensively investigate how to design a framework for human mesh recovery from arbitrary multi-view images and conduct extensive experiments to validate the effect of proposed components. To demonstrate the superiority of the proposed method, the comparison with state-of-the-art (SOTA) methods is performed on all three datasets.   

Human3.6M is a large 3D human pose benchmark, containing 11 different subjects and each subject with 15 different daily indoor actions. The subjects are captured by 4 synchronized 50 Hz digital cameras. The dataset provides 2D/3D keypoint annotations obtained by marker-based mocap system. The SMPL annotations are obtained by applying MoSh~\cite{loper2014mosh} to the sparse 3D MoCap marker following previous works~\cite{kanazawa2018hmr, kolotouros2019graphcmr}.
Similar to \cite{kanazawa2018hmr,liang2019mv-shapeaware, shin2020mv-lva}, we follow the protocol 1 to use subjects 1, 5, 6, 7, and 8 for training and subjects 9 and 11 are for testing.

% We use subjects 9 and 11 in protocol 1 for model learning and validation.

MPI-INF-3DHP contains 8 actors 
% (4 males and 4 females) 
performing 8 human activities. 2D and 3D keypoint annotations are also provided by marker-less mocap system. The dataset contains multi-view images of 14 cameras with regular lenses, and we choose views 0, 2, 7, 8 from them. Following~\cite{liang2019mv-shapeaware, shin2020mv-lva}, the standard test dataset of MPI-INF-3DHP simply includes single-view images, we use subjects 1-7 for training and subject 8 for testing.

TotalCapture dataset consists of 1.9 million frames captured from 8 calibrated full HD video cameras recording at 60Hz.
% The images of “Freestyle3”, “Walking2” and “Acting3” on subjects 1,2,3,4 and 5 are used for our experiments. In addition, these images from four cameras (1,3,5,7) in experiments and are without the IMU sensors.
As suggested in~\cite{trumble2017totalcapture}, the training set consists of “ROM1,2,3”, “Walking1,3”, “Freestyle1,2”, “Acting1,2”, “Running1” on subjects 1, 2, and 3. The testing set consists of “Freestyle3”, “Walking2” and “Acting3” on subjects 1, 2, 3, 4, and 5. We use the images from four cameras (1, 3, 5, 7) in our experiments and do not use the IMU sensors.

\noindent
{\bf Evaluation Metrics.}
For Human3.6M dataset, we use mean per joint position error (MPJPE) and reconstruction error as metrics. Reconstruction error is MPJPE after rigid alignment of the prediction with ground truth via Procrustes. The reconstrution error is also known as PA (procrustes aligned)-MPJPE. For MPI-INF-3DHP, as this dataset is collected indoors and outdoors with a multi-camera marker-less MoCap system, thus the 3D annotations are less accurate. So in addition to MPJPE and PA-MPJPE, the Percentage of Correct Keypoints (PCK) thresholded at 150 mm and the Area Under the Curve (AUC) over a range of PCK thresholds with interval of 5 mm before and after rigid alignment are both reported on this dataset. For TotalCapture dataset, we report MPJPE and PA-MPJPE across different subjects and activity sequences.
% of different subjects and different action sequences as in~\cite{trumble2017totalcapture}.

\noindent
{\bf Implementation Details.}
In our implementation, input size of all images is $256\times256$. For the 2D image encoder, we use the ResNet-50~\cite{he2016resnet} as the standard backbone following~\cite{kanazawa2018hmr, li2021mv-spin, kolotouros2021prohmr}. However, as depicted in~\cite{goel20234dhumans}, using a ViT backbone~\cite{dosovitskiy2020vit} achieves a significant improvement across the 3D and 2D metrics. Thus, in addition to ResNet-50, we also conduct experiments using a ViT backbone. As we use a combination of multiple losses, during training, we weight the different loss terms empirically as~\cite{goel20234dhumans}. The model is trained for 100 epochs on 2 NVIDIA A100 GPUs and we use the AdamW optimizer with a learning rate of $1e-5$, $\beta_1 = 0.9$, $\beta_2 = 0.999$, and a weight decay of $1e-4$.
\begin{figure}[t]
    \centering
    \includegraphics[width=\linewidth]{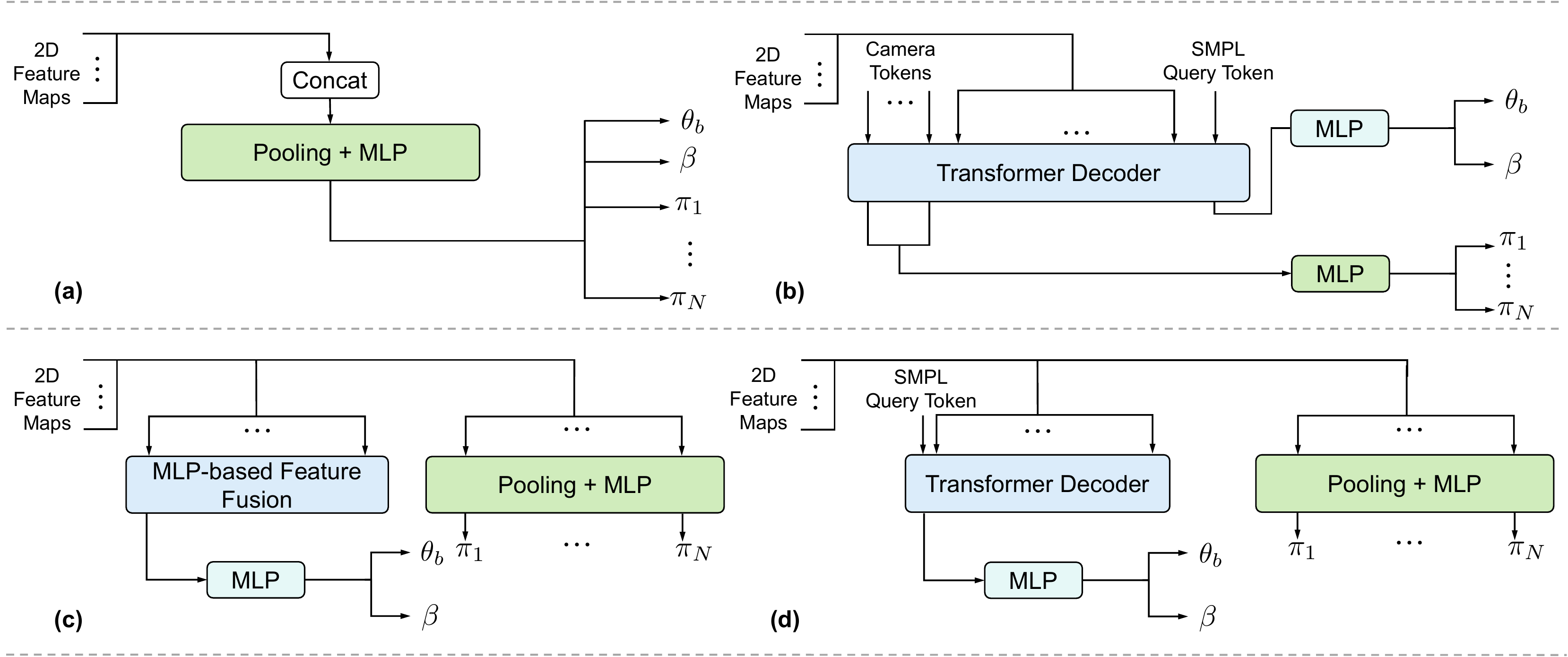}
    \caption{{\bf Different architecture designs for human mesh recovery from multi-view images.} The 2D feature maps are produced by the same 2D image encoder. All these variants output human mesh parameters and camera parameters of all views.}
    \label{fig:ablation_design}
\end{figure}
\subsection{Ablation Studies}

\noindent
{\bf Design of Decoupling and Multi-view Fusion.}
We systematically investigate how to design a concise and unified network architecture to efficiently and effectively accommodate randomly varied camera poses and number of views. We design four different architectures for human mesh recovery from multi-view images, as shown in \cref{fig:ablation_design}, we compare them both theoretically and empirically. {\bf (a)} This is a simple extension of single-view HMR~\cite{kanazawa2018hmr}. The 2D feature maps from 
different-view images are concatenated and pooled along two spatial dimensions (width and height),
% \fnote{how the pooling operates on the concatenated feature.}
% \xnote{the feature maps product by backbone is $C\times H \times W$ for each view, concatenated to $ (N\times C)\times H \times W$, the pooling is operating on the last two dimension and produce feature vector of $N\times C$-dimension, while is $C$-dimension for single-view}
producing a single feature vector whose dimension is dependent on the number of views. Then the feature vector is forwarded to the MLP to output all parameters, including pose/shape parameters and camera parameters of all views. This architecture is concise but not flexible as the dimension of feature vector, \ie, the dimension of the layers in the MLP, is dependent on the number of views. Moreover, the output dimension of the MLP is related with the number of camera pose as well, since this architecture produces the parameters all together. Thus it cannot be directly adapted to different number of camera views.
% \fnote{the MLP is also related with the number of camera pose to be estimated.}
% \xnote{Yes, but the number of camera pose to be estimated is exactly the number of views, `related with the number of camera pose to be estimated' is from the perspective of output, right? }
{\bf (b)} This architecture decouples camera parameters and body parameters to some extent. Camera parameters of different views and body parameters are represented as independent query tokens, which are used to produce corresponding parameters after processing in the transformer decoder. Although the estimation of body parameters is independent on the number of views, the number of camera query tokens still depend on the number of views, making this architecture unable to cope with an arbitrary number of views. {\bf (c)} This architecture is similar to our final method while using an MLP
% \fnote{Why "an MLP"?} \xnote{Because in the case of abbreviations, the first syllable of the letter M is a vowel `e'}
for feature fusion instead of a transformer decoder. Given 2D feature maps of different views, unlike {\bf (a)}, where the concatenating and pooling are employed to produce feature vector, we use an extra MLP to produce attention scores, \ie, weights for different views,
% \fnote{can the number of weights change with the number of views?}
% \xnote{Yes, because the MLP is also shared across views, it produce attention scores for each view}
then the attention scores are normalized and used to calculate the weighted sum of features of different views to produce a fused feature vector, this fusion method is similar to the fusion of global features in~\cite{yu2022mv-uncalibrated}. This method is independent on the number of views, and has the same flexibility as our final method but with less feature representation capacity.
% \fnote{what are the drawbacks of this architecture?}
% \xnote{It does not use transformer decoder as we do, thus not fuse multi-view information effectively}
% \fnote{so in this architecture, the attention is view level not region level, right?}
% \xnote{I am not sure, the attention scores here are just a weight vector (produced by MLP) whose shape are same as the feature vector, and are normalized to weight features from multi-view}
{\bf (d)} Our final method decouples the estimation of pose/shape and camera parameters, and uses a transformer decoder to fuse features from multi-view images.

\begin{table}[t]
    \centering
    \caption{{\bf Ablation study of different architecture designs.} To show the effect of different variants as much as possible, we use the powerful ViT as the 2D image encoder for all variants and report MPJPE and PA-MPJPE on Human3.6M dataset.}
    \vspace{-0.5em}
    \resizebox{0.7\linewidth}{!}{
        \begin{tabular}{l|cc}
            \bottomrule
            Variant                                 & MPJPE~$\downarrow$ & PA-MPJPE~$\downarrow$ \\ \hline
            (a) MLP                                 & \tbest{31.4}       & \tbest{25.3}          \\
            (b) Independent tokens                  & \best{29.8}        & \sbest{23.2}          \\
            (c) Decoupling with MLP                 & 36.9               & \tbest{25.3}          \\
            (d) Decoupling with Transformer decoder & \sbest{31.0}       & \best{22.8}           \\ \toprule
        \end{tabular}
    }
    \vspace{-1em}
    \label{tab:ablation_design}
\end{table}

The results of different architecture designs are presented in \cref{tab:ablation_design}. We use the same 2D image encoder (\ie, ViT) for the fair of comparison. Compared to other designs, our final method achieves best PA-MPJPE and second best MPJPE, and are adaptive to arbitrary multi-view scenario,
which demonstrates the superiority of our method. In contrast to methods directly estimating parameters using MLP {\bf (a)}, and fusing features from multi-view images using MLP {\bf (c)}, our method outperforms them notably. It is worth noting that the performance of {\bf (b)} is competitive to our final method {\bf (d)},
% \fnote{the MPJPE of "c" is 36.9, "d" is 31, c is not slightly lower than ours.}, 
% \xnote{should be `b'}
and we consider this result can be attributed to three perspectives. First, both of these two designs adopt transformer decoder to aggregate features from multi-view images and achieve top-two results on both PA-MPJPE and MPJPE, demonstrating the powerful capability of the transformer decoder for features fusion from multi-view images. Second, compared to PA-MPJPE, MPJPE indicates the error before rigid alignment, thus is able to demonstrate the accuracy of camera parameters estimation to some extent. In {\bf (b)} the the camera tokens also pass through the transformer decoder to produce camera parameters while in {\bf (d)} the camera parameters are simply produced by the MLP from corresponding features. The intuitive reason is that the former method uses stronger features to predict camera parameters, thus has slightly lower MPJPE. Third, although {\bf (b)} reports slightly lower MPJPE, it is less flexible than our final method. In other words, by a small compromise of the performance indicated by MPJPE, our final method provides a higher degree of flexibility, so that our method achieves better trade-off between performance and flexibility.

\noindent
{\bf Number of Views.}
\begin{table}[t]
\begin{minipage}{0.49\textwidth}
    \centering
    \caption{{\bf Ablation study and comparison results of number of views on Human3.6M dataset.} 
    % Note that PaFF~\cite{jia2023mv-paff} uses camera calibrations for fusion, and the results reported here are from the models re-trained using 2 and 3 views respectively, while Yu~\etal~\cite{yu2022mv-uncalibrated} and our method require neither camera calibrations nor retraining, so the results of PaFF are expected to provide the performance upper bound. We use ResNet-50 as the 2D image encoder for our method as the other two methods do for the fair of comparison.
    }
    \resizebox{\linewidth}{!}{
    \renewcommand\arraystretch{1.2}
        \begin{tabular}{c|c|cccc}
        
            \bottomrule
            \multirow{2}{*}{Method}                                & \multirow{2}{*}{Metrics} & \multicolumn{4}{c}{Number of views}                      \\ \cline{3-6}
                                                                   &                          & 4                                   & 3    & 2    & 1    \\ \hline
            \multirow{2}{*}{PaFF~\cite{jia2023mv-paff}}            & MPJPE~$\downarrow$       & \best{33.0}                                & \best{33.5} & \best{33.8} & -    \\ \cline{2-6}
                                                                   & PA-MPJPE~$\downarrow$    & \best{26.9}                                & \best{27.5} & \best{27.6} & -    \\ \hline
            \multirow{2}{*}{Yu~\etal~\cite{yu2022mv-uncalibrated}} & MPJPE~$\downarrow$       & -                                   & -    & -    & -    \\ \cline{2-6}
                                                                   & PA-MPJPE~$\downarrow$    & \tbest{33.0}                                & \tbest{34.2} & \tbest{37.3} & \sbest{44.1} \\ \hline
            \multirow{2}{*}{Ours}                                  & MPJPE~$\downarrow$       & \sbest{36.3 }                               &   \sbest{38.7}   & \sbest{ 42.5}    &   \best{52.9}   \\ \cline{2-6}
                                                                   & PA-MPJPE~$\downarrow$    & \sbest{28.3}                                &   \sbest{30.9}   &  \sbest{34.5}    &  \best{42.5}    \\
            \toprule
        \end{tabular}
    \label{tab:ablation_num_of_views}
    }
\end{minipage}
\vspace{-1em}
\begin{minipage}{0.49\textwidth}
    \centering
    \captionof{figure}{{\bf The performance trend w.r.t the number of views.}}
    \includegraphics[width=0.83\linewidth]{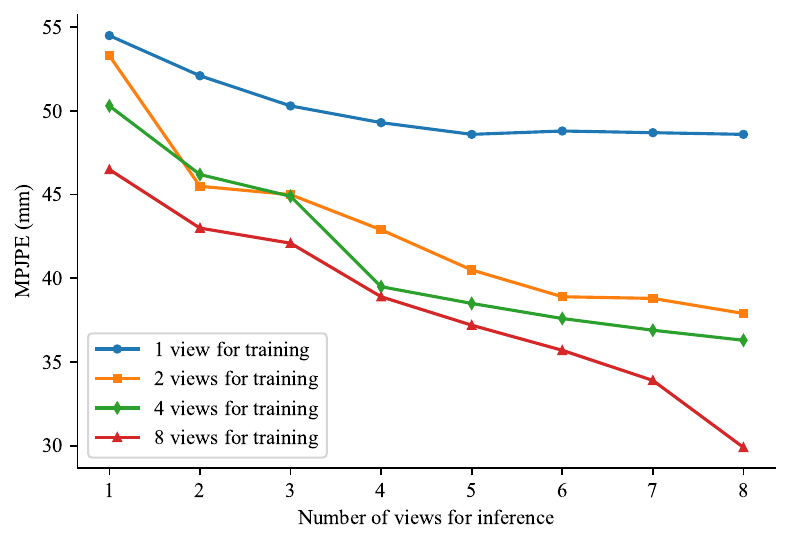}
    \label{fig:trend}
\end{minipage}
\vspace{-1em}
\end{table}
As we claim that our method can reconstruct human pose and shape for arbitrary multi-view images, to validate the flexibility, \ie, the ability to adapt to different number of views, we conduct ablation study on the effect of varying number of views during inference phase.
% \fnote{how is the model trained? Is the model trained by only 4-view or a combination of 1-view, 2-view,3-view, and 4-view?}
% \xnote{Yes, the model is retrained using the latter way as Yu~\etal~\cite{yu2022mv-uncalibrated} did.}
Since there have been two methods, Yu~\etal~\cite{yu2022mv-uncalibrated} and PaFF~\cite{jia2023mv-paff}, state that they can be applied to arbitrary number of views, we compare with them on Human 3.6M dataset and the results are shown in \cref{tab:ablation_num_of_views}. For fair comparison, our model is trained with a combination of varying number of multi-view image (1-view, 2-view, 3-view, and 4-view), which is same as ~\cite{yu2022mv-uncalibrated} does. While PaFF 
claims to have the ability to be applied to different number of views, the reported results are from the model retrained and tested using same number of views.
% \fnote{a little confused, does PaFF use same multi-view in both training and test?}
% \xnote{yes, for each view number, the model is retrained and tested}
Moreover, PaFF requires camera calibrations for feature fusion. So the results of PaFF can be expected to provide the performance upper bound of other two methods which are calibration-free and not retrained. Compared with Yu~\etal~\cite{yu2022mv-uncalibrated}, our method achieves better performance at any number of views.

We adopt varying number of views for training (1, 2, 4, 8) and inference (1-8) on MPI-INF-3DHP dataset. The performance trend w.r.t the number of views is in the Fig.~\ref{fig:trend}.
% We can observe that using more views for inference achieve better performance as more multi-view information is provided, and using more views for training also helps a lot as model will better learn to fuse multi-view information.
Note that in both training and inference phases the model precision grows with the increasing number of views and our model achieves reasonable performance when \# training views $>$ \# inference views or vice versa.  

\subsection{Comparison with State-of-the-art Methods}
\begin{table}[t]
    \begin{minipage}{0.49\textwidth} % Create a minipage occupying half of the page width
        \centering
    %     \caption{{\bf Comparison results on Human3.6M dataset.} We compare our method with single-view methods, calibration-requiring multi-view methods and calibration-free multi-view methods. $^\dag$ denotes the output of methods are non-parametric.
    % % \fnote{may need gray display on ours single view results.}
    % }
     \caption{{\bf Comparison results on Human3.6M dataset.} We compare our method with single-view, calibration-requiring multi-view, and calibration-free multi-view methods. $^\dag$ denotes the output of methods are non-parametric.
    % \fnote{may need gray display on ours single view results.}
    }
    \resizebox{\linewidth}{!}{
        \begin{tabular}{c|cc|cc}
            % \toprule
            \bottomrule
            Method                                     & Multi-view & Calibration-free & MPJPE~$\downarrow$ & PA-MPJPE~$\downarrow$ \\
            % \midrule
            \hline
            SMPLify~\cite{bogo2016smplify}             & No         & -                & -                  & 82.3                  \\
            Pavlakos~\etal~\cite{pavlakos2018learning} & No         & -                & -                  & 75.9                  \\
            HMR~\cite{kanazawa2018hmr}                 & No         & -                & 88.0               & 56.8                  \\
            NBF~\cite{omran2018nbf}                    & No         & -                & -                  & 59.9                  \\
            GraphCMR~\cite{kolotouros2019graphcmr}$^\dag$     & No         & -                & -                  & 50.1                  \\
            HoloPose~\cite{guler2019holopose}          & No         & -                & 60.3               & 46.5                  \\
           DenseRac~\cite{xu2019denserac}             & No         & -                & 76.8               & 48.0                  \\
            SPIN~\cite{kolotouros2019spin}             & No         & -                & 62.5               & 41.1                  \\
            DaNet~\cite{zhang2019danet}                & No         & -                & 61.5               & 48.6                  \\
            DecoMR~\cite{zeng2020decomr}$^\dag$               & No         & -                & -                  & 39.3                  \\
            LearnedGD~\cite{song2020learnedGD}         & No         & -                & -                  & 56.4                  \\
            Pose2Mesh~\cite{choi2020pose2mesh}$^\dag$         & No         & -                & 64.9               & 47.0                  \\
            HKMR~\cite{georgakis2020hkmr}              & No         & -                & 59.6               & 43.2                  \\
            I2L-MeshNet~\cite{moon2020i2l-meshnet}$^\dag$     & No         & -                & 55.7               & 41.1                  \\
            HybrIK~\cite{li2021hybrik}                 & No         & -                & 33.6               & 55.4                  \\
            METRO~\cite{lin2021metro}$^\dag$                  & No         & -                & 54.0               & 36.7                  \\
            ProHMR~\cite{kolotouros2021prohmr}         & No         & -                & -                  & 41.2                  \\
            Graphormer~\cite{lin2021metro}$^\dag$             & No         & -                & 51.2               & 34.5                  \\
            PyMAF~\cite{zhang2021pymaf}                & No         & -                & 57.7               & 40.5                  \\
            PARE~\cite{kocabas2021pare}                & No         & -                & 76.8               & 50.6                  \\
            CLIFF~\cite{li2022cliff}                   & No         & -                & 47.1               & 32.7                  \\
            FastMETRO~\cite{cho2022fastmetro}$^\dag$          & No         & -                & 52.2               & 33.7                  \\
            PyMAF-X~\cite{zhang2023pymafx}             & No         & -                & 54.2               & 37.2                  \\
            POTTER~\cite{zheng2023potter}              & No         & -                & 56.5               & 35.1                  \\
            FeatER~\cite{zheng2023feater}              & No         & -                & 49.9               & 32.8                  \\
            ProPose~\cite{fang2023propose}             & No         & -                & 45.7               & 29.1                  \\
            PLIKS~\cite{shetty2023pliks}               & No         & -                & 47.0               & 34.5                  \\
            PointHMR~\cite{kim2023sampling}$^\dag$            & No         & -                & 48.3               & 32.9                  \\
            VirtualMarker~\cite{ma2023virtualmarkers}  & No         & -                & 47.3               & 32.0                  \\
            HMR 2.0a~\cite{goel20234dhumans}           & No         & -                & 44.8               & 33.6                  \\
            HMR 2.0b~\cite{goel20234dhumans}           & No         & -                & 50.0               & 32.4                  \\
            Yu~\etal~\cite{yu2022mv-uncalibrated}      & No        & -              & -                  & 44.1                  \\
            Ours (SV, ResNet-50)                       & No         & -                &    {\cellcolor{LightGrey}52.9}                &           {\cellcolor{LightGrey}42.5}            \\
            Ours (SV, ViT)                             & No         & -                &      {\cellcolor{LightGrey}43.3}              &           {\cellcolor{LightGrey}32.6}            \\
            % \midrule
            \hline
            % MuVS~\cite{huang2017towards}               & Yes        & No               & 58.2               & 47.1                  \\
            MV-SPIN~\cite{shin2020mv-lva}              & Yes        & No               & 49.8               & 35.4                  \\
            LVS~\cite{shin2020mv-lva}                  & Yes        & No               & 46.9               & 32.5                  \\
            PaFF~\cite{jia2023mv-paff}                 & Yes        & No               & 33.0               & 26.9                  \\
            LMT~\cite{chun2023mv-lmt}                 & Yes        & No               & 30.1               & -                  \\
            % \midrule
            \hline
            Liang~\etal~\cite{liang2019mv-shapeaware}  & Yes        & Yes              & 79.9               & 45.1                  \\
            ProHMR~\cite{kolotouros2021prohmr}         & Yes        & Yes              & 62.2               & 34.5                  \\
            Yu~\etal~\cite{yu2022mv-uncalibrated}      & Yes        & Yes              & -                  & 33.0                  \\
            Calib-free PaFF~\cite{jia2023mv-paff}      & Yes        & Yes              & \tbest{44.8}       & \sbest{28.2}          \\
            Ours (ResNet-50)                           & Yes        & Yes              & \sbest{36.3}       & \tbest{28.3}          \\
            Ours (ViT)                                 & Yes        & Yes              & \best{31.0}        & \best{22.8}           \\
            % \bottomrule
            \toprule
        \end{tabular}
        \label{tab:comparison_h36m}
    }
    \end{minipage}
    \vspace{-1em}
    \begin{minipage}{0.49\textwidth} % Create another minipage occupying half of the page width
        \centering
        \caption{{\bf Comparison results on MPI-INF-3DHP dataset.} The methods are categorized from top to bottom as single-view, calibration-requiring multi-view, and calibration-free multi-view methods.}
    \resizebox{\linewidth}{!}{
        \begin{tabular}{c|ccc|ccc}
            \bottomrule
            \multirow{2}{*}{Method}                   & \multicolumn{3}{c|}{Absolute} & \multicolumn{3}{c}{Rigid Alignment}                                                                           \\
            \cline{2-7}
                                                      & PCK $\uparrow$                & AUC$\uparrow$                       & MPJPE$\downarrow$ & PCK $\uparrow$ & AUC $\uparrow$ & MPJPE$\downarrow$ \\
            \hline
            % VNect~\cite{mehta2017vnect}        & 76.6           & 40.0           & 124.7        & 83.9                          & 47.3                                & 98.0                          \\
            HMR~\cite{kanazawa2018hmr}       & 72.9           & 36.5           & 124.2         & 86.3                          & 47.8                                & 89.8                           \\
            SPIN~\cite{kolotouros2019spin}            & 76.4                          & 37.1                                & 105.2             & 92.5          & 55.6           & 67.5             \\
            ProHMR~\cite{kolotouros2021prohmr}        & -                             & -                                   & -                 & -              & -              & 65.0              \\
            Ours (SV, ResNet-50)                      &   {\cellcolor{LightGrey}71.3}                            &           {\cellcolor{LightGrey}24.3}                          &         {\cellcolor{LightGrey}77.8}          &        {\cellcolor{LightGrey}94.3}        &      {\cellcolor{LightGrey}50.0}          &  {\cellcolor{LightGrey}64.5  }               \\
            Ours (SV, ViT)                            &     {\cellcolor{LightGrey} 99.2 }                       &         {\cellcolor{LightGrey}66.1 }                           &        {\cellcolor{LightGrey} 54.5 }         &        {\cellcolor{LightGrey} 99.3 }      &      {\cellcolor{LightGrey} 73.3 }        & {\cellcolor{LightGrey} 43.7    }             \\
            \hline
            LVS~\cite{shin2020mv-lva}                 & -                             & -                                   & -                 & 97.4           & 65.5           & 50.2              \\
            PaFF~\cite{jia2023mv-paff}                & -                             & -                                   & -                 & 98.6           & 67.3           & 48.4              \\
            LMT~\cite{chun2023mv-lmt}                & -                             & -                                   & -                 & 96.6           & 71.6           & 45.9              \\
            \hline
            Liang~\etal~\cite{liang2019mv-shapeaware} & \tbest{72.0}                          & \tbest{35.0}                                & \tbest{126.0}             & \tbest{95.0}           & \tbest{65.0}           & \tbest{59.0}              \\
            Ours (ResNet-50)                          &     \sbest{94.0}                          &                  \sbest{44.0}                   &         \sbest{55.4}          &         \sbest{98.8}       &       \sbest{71.0}         & \sbest{39.8}                  \\
            Ours (ViT)                                &       \best{99.4}                        &           \best{74.0}                          &  \best{39.7}                 &     \best{99.9}           &       \best{81.8}         &          \best{29.2}         \\
            \toprule
        \end{tabular}
        \label{tab:comparison_mpii3d}
    }
% \caption{{\bf Comparison results on TotalCapture dataset.} Different methods use input information from different sensors provided by the dataset.}
\caption{{\bf Comparison results on TotalCapture dataset.} We distinguish different methods by the input information from different sensors (Cameras and /or IMUs) provided by the dataset they use.}
    \resizebox{\linewidth}{!}{
        \begin{tabular}{c|c|cc}
            % \toprule
            \bottomrule
            Method                                     & Sensors  & MPJPE~$\downarrow$ & PA-MPJPE~$\downarrow$ \\
            % \midrule
            \hline
            VIP~\cite{marcard2018vip}                                   &     SV+IMUs             &            -        &           26.0            \\
                    Kalman Filter~\cite{fang2023propose}                                   &     SV+IMUs            &        34.7            &     23.1                  \\
                    ProPose~\cite{fang2023propose}                                   &      SV+IMUs             &           28.5         &          21.2             \\
                   ProPose~\cite{fang2023propose}                                   &      SV              &           42.1         &          29.0             \\
                    Ours (ResNet-50)                       & SV                     &             {\cellcolor{LightGrey} 41.4 }     &            {\cellcolor{LightGrey} 28.9  }        \\
            Ours (ViT)                             & SV                  &    {\cellcolor{LightGrey}  38.5 }             &       {\cellcolor{LightGrey}   27.2 }            \\
            \hline
            PVH~\cite{trumble2017totalcapture}                                   &     MV             &          107.3          &            -           \\
                    Tri-CPM~\cite{wei2016cpm}                                   &      MV           &           99.8         &         -              \\
            IMUPVH~\cite{gilbert2019pvh}                                   &     MV+IMUs             &          42.6          &             -          \\
                    GeoFuse~\cite{zhang2020geofuse}                                   &     MV+IMUs             &         24.6           &          20.6             \\
                     ProPose~\cite{fang2023propose}                                   &     MV+IMUs            &        23.5            &     19.4                  \\
                    ProHMR~\cite{kolotouros2021prohmr}                                   &    MV              &            127.8        &              -         \\
                    Trumble~\etal~\cite{trumble2018deep}                                   &    MV               &           \tbest{85.4}         &          -             \\
                    
            Ours (ResNet-50)                       & MV                        &    \sbest{36.4}                &           \sbest{26.1}            \\
            Ours (ViT)                             & MV                    &  \best{29.2}                  &     \best{21.2}                  \\
            \toprule
        \end{tabular}
        \label{tab:comparison_totalcapture}
    }
    \end{minipage}
    \vspace{-1em}
\end{table}

\noindent
{\bf Human3.6M.}
The comparison results on Human3.6M dataset are shown in \cref{tab:comparison_h36m}. As our method is suitable for arbitrary multi-view images including single-view image, we also report the results of our method for single-view human body reconstruction as Yu~\etal~\cite{yu2022mv-uncalibrated} do, which provides a fair comparison with other single-view methods. We report the results of method using ResNet-50 and ViT as 2D image encoder. For single-view human body reconstruction, our method achieves comparable results to latest SOTA method, \ie, HMR 2.0, and achieves better results than Yu~\etal~\cite{yu2022mv-uncalibrated}. As a calibration-free multi-view method, our method also achieves SOTA results which are comparable to, or even better than calibration-requiring methods.

\noindent
{\bf MPI-INF-3DHP.}
The comparison results on MPI-INF-3DHP dataset are listed in \cref{tab:comparison_mpii3d}. As the dataset is collected using marker-less MoCap system, the ground truth 3D annotations have some noise and is less accurate, thus PCK and AUC are reported in addition to MPJPE, the results before and after rigid alignment are reported. We compare our method with single-view methods, calibration-requiring multi-view methods, and calibration-free multi-view methods. For reconstructing human body from multi-view images, our method outperforms both calibration-requiring and calibration-free methods. And our results of single-view reconstruction are also superior to the results from other single-view methods.

\noindent
{\bf TotalCapture.}
\Cref{tab:comparison_totalcapture} displays the comparison results on TotalCapture dataset. In addition to RGB cameras, the dataset also provides information from IMU sensors and some methods use the information as the input. So we distinguish different methods by the input information from different sensors (cameras and /or IMUs) provided by the dataset they use.

\subsection{Qualitative Comparison}
\begin{figure}
    \centering
    \includegraphics[width=\linewidth]{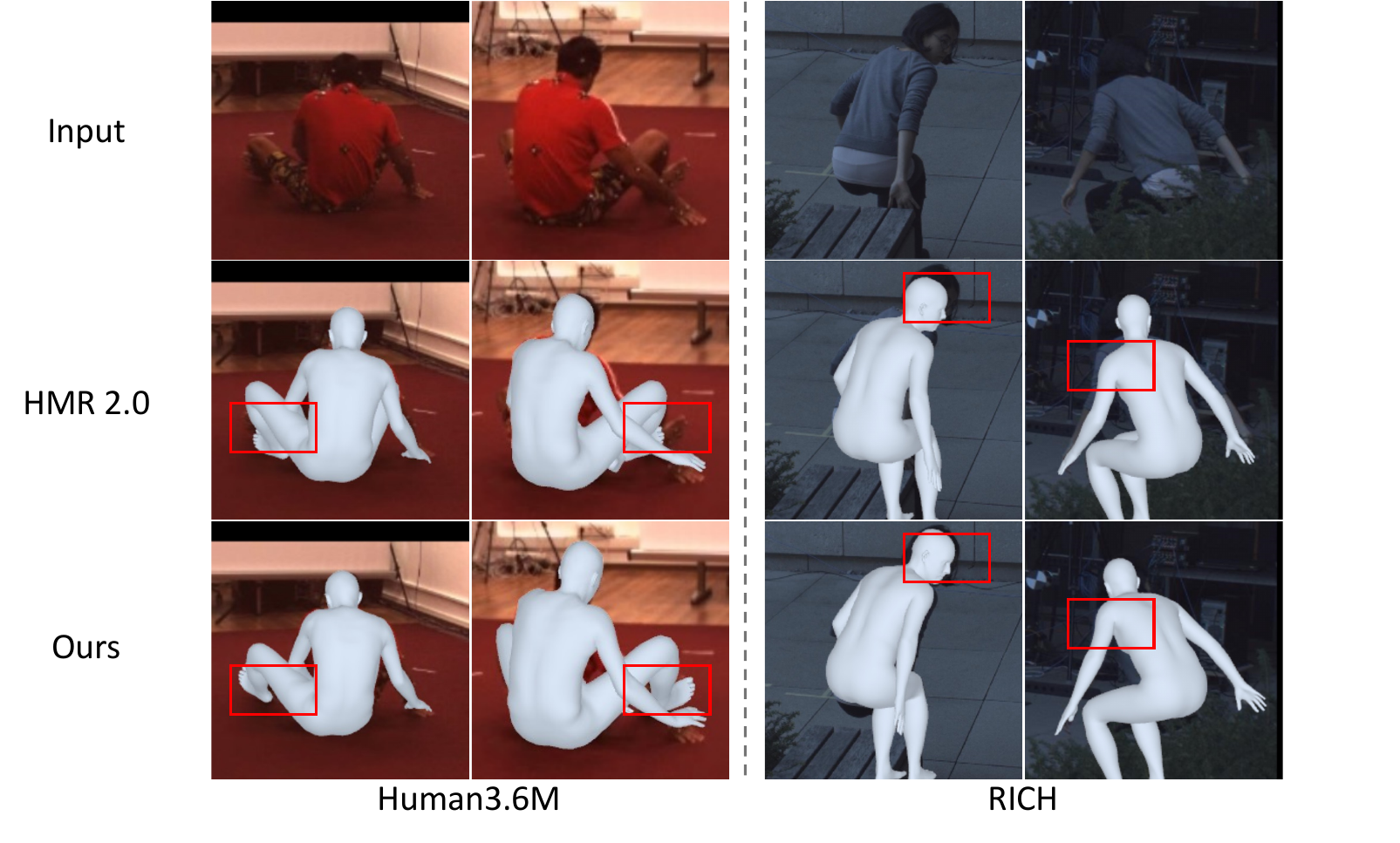}
    \caption{\bf Qualitative comparison results with HMR2.0~\cite{goel20234dhumans}.}
    \label{fig:quali_eval}
\end{figure}

We show the results of  comparison with HMR 2.0~\cite{goel20234dhumans} using images from indoor dataset, Human3.6M, and in-the-wild dataset, RICH, in the Fig.~\ref{fig:quali_eval}. It can be observed that our method provides more accurate body mesh (as shown in red boxes), and more importantly, the multi-view consistency. 
\section{Conclusion}
\label{sec:conclusion}
We propose a simple, flexible, and effective framework using a divide and conquer strategy for human mesh recovery from arbitrary multi-view images. Based on the observation that camera poses and human mesh are independent of each other, we split the estimation of them into two sub-tasks. For the estimation of camera poses, we use a single shared MLP to process all views in a parallel way. For the estimation of human mesh, to fuse multi-view information effectively and make the fusion independent of the number of views, we introduce a transformer decoder with a SMPL query token to extract cross-view features for mesh recovery. The framework is not only adaptive to arbitrary multi-view images but also considerable effective in multi-view information fusion. 
% Our framework can be easily adopted to different camera settings, thus benefits further researches.
% , \eg, building an universal 3D body modeling network.

\clearpage
% ---- Bibliography ----
%
% BibTeX users should specify bibliography style 'splncs04'.
% References will then be sorted and formatted in the correct style.
%
\bibliographystyle{splncs04}
\bibliography{main}
\appendix
\clearpage
\section{Computational Overhead}
To validate the efficiency, we analyze the computational overhead of our method. We report the number of parameters, multiply-add operations (MACs), and time consumption per sample of different number of views in \cref{tab:computational overhead}. For clearer illustration, we divide the whole model into two parts, the backbone (ViT), 
for feature extraction 
and the head (both CPE and AVF) for camera and human body mesh parameters estimation.
% \fnote{replace fusion with disentangle?, to make it clear that our method is not to fuse feature}
% \fnote{AVF? or AVF\&CBD?}
 The results show that the head adds very little parameters and computational overhead compared to the ViT backbone, which means that our method is efficient.
% \begin{table}[htbp]
% \centering
% \caption{{\bf Computational overhead}}
% \begin{tabular}{@{}c|c|c|c|c|c|c@{}}
% \bottomrule

% \multirow{2}{*}{\makecell{Number of \\ Views}}& \multicolumn{2}{c|}{Parameters (M)}    & \multicolumn{2}{c|}{MACs (G)}& \multicolumn{2}{c}{Time (ms)} \\ \cline{2-7}
% & Backbone&Head&Backbone&Head&Backbone&Head\\
%      % &           & Module     &    & Module  & & Module  \\
%      \hline
%    1       & \multirow{4}{*}{630.7} & \multirow{4}{*}{60.1} &  121.0          &    4.7    & 55.2&   10.1      \\
%    2       &                   &                   &    242.1        &         9.4   & 107.9&  12.6   \\
%     3      &                   &                   &      363.1      &          14.1   &148.8& 12.7   \\
%     4      &                   &                   &      484.2      &          18.8     &181.1&  13.9\\
% \toprule
% \end{tabular}
% \label{tab:computational overhead}
% % }
% % \caption{Computational overhead}
% \end{table}
\section{Sensitivity to Input}
Although we use ground-truth bounding boxs of human in our experiments, in real
applications, there would be imperfect detection/crop such as scale changes and center shifts. To investigate the effect of inaccurate detection/crop, we perturb the input images from Human3.6M, \ie, 
add random center shifts ($\pm$~20 pixels)
% add random shifts ($\pm$~20 pixels) to the bbox center
% \fnote{how large the shift is} 
and randomly change the scales (0.8-1.2),
% \fnote{how much the scale is changed}, 
then test using the perturbed images without retraining, the results are in \cref{tab:sensitive to the input}. 
% The results are listed below.
We observe that imperfect
% detection/
crop does affect performance, but the performance degradation is acceptable, especially with ViT as backbone. On one hand, ViT is stronger 
% than ResNet 
and produces more robust features, and on the other hand, our method could fuse information from different views effectively thus the imperfections of the crops
% input in some views 
could be suppressed.
% \begin{table}[htbp]
% \centering
% \caption{Sensitive to the input}
% % \vspace{-1em}
% % \resizebox{0.95\linewidth}{!}{
% \begin{tabular}{@{}c|c|c|c|c@{}}
% \bottomrule

% \multirow{2}{*}{Method} & \multicolumn{2}{c|}{ResNet-50}    & \multicolumn{2}{c}{ViT} \\ \cline{2-5}
%      & MPJPE          & PA-MPJPE     & MPJPE          & PA-MPJPE   \\ \hline
%   Ours        & 36.3 & 28.3 &       31.0     &         22.8       \\
%    Ours (Perturbed)       &     59.5              &       39.4            &  46.8          &    30.1           \\

% \toprule
% \end{tabular}
% % }
% % \caption{Perturb}
% \end{table}

\begin{table}[h]
    \begin{minipage}{0.49\textwidth}
        \centering
        \caption{{\bf Computational overhead} 
        Efficiency analysis of the method using different number of views.
        }
        \resizebox{0.94\linewidth}{!}{
\begin{tabular}{@{}c|c|c|c|c|c|c@{}}
\bottomrule

\multirow{2}{*}{\makecell{Number of \\ Views}}& \multicolumn{2}{c|}{Parameters (M)}    & \multicolumn{2}{c|}{MACs (G)}& \multicolumn{2}{c}{Time (ms)} \\ \cline{2-7}
& Backbone&Head&Backbone&Head&Backbone&Head\\
     % &           & Module     &    & Module  & & Module  \\
     \hline
   1       & \multirow{4}{*}{630.7} & \multirow{4}{*}{60.1} &  121.0          &    4.7    & 55.2&   10.1      \\
   2       &                   &                   &    242.1        &         9.4   & 107.9&  12.6   \\
    3      &                   &                   &      363.1      &          14.1   &148.8& 12.7   \\
    4      &                   &                   &      484.2      &          18.8     &181.1&  13.9\\
\toprule
\end{tabular}
\label{tab:computational overhead}
}
    \end{minipage}
    \begin{minipage}{0.49\textwidth}
        \centering
        % \caption{{\bf Sensitivity to input.} We perturb the center and scale of ground truth bbox in Human3.6M dataset and test model using the perturbed images without retraining.}
        \caption{{\bf Sensitivity to input.} We perturb the 
        input images
        % centers and scales of ground truth bboxs 
        of Human3.6M dataset and test model without retraining.}
% \vspace{-1em}
\resizebox{\linewidth}{!}{
\begin{tabular}{@{}c|c|c|c|c@{}}
\bottomrule

\multirow{2}{*}{Method} & \multicolumn{2}{c|}{ResNet-50}    & \multicolumn{2}{c}{ViT} \\ \cline{2-5}
     & MPJPE          & PA-MPJPE     & MPJPE          & PA-MPJPE   \\ \hline
  Ours        & 36.3 & 28.3 &       31.0     &         22.8       
  \\Ours (Scale changing)       &     47.2              &       34.1            &  38.9          &    25.9           \\
  Ours (Center shifts)       &     52.6              &       35.6            &  41.7          &    26.7           \\
   Ours (Perturbed)       &     59.5              &       39.4            &  46.8          &    30.1           \\

\toprule
\end{tabular}
\label{tab:sensitive to the input}
}
    \end{minipage}
\end{table}
\section{Images from Unseen Views} \label{sec:supp_1}
\begin{figure*}
    \centering
    \includegraphics[width=\linewidth]{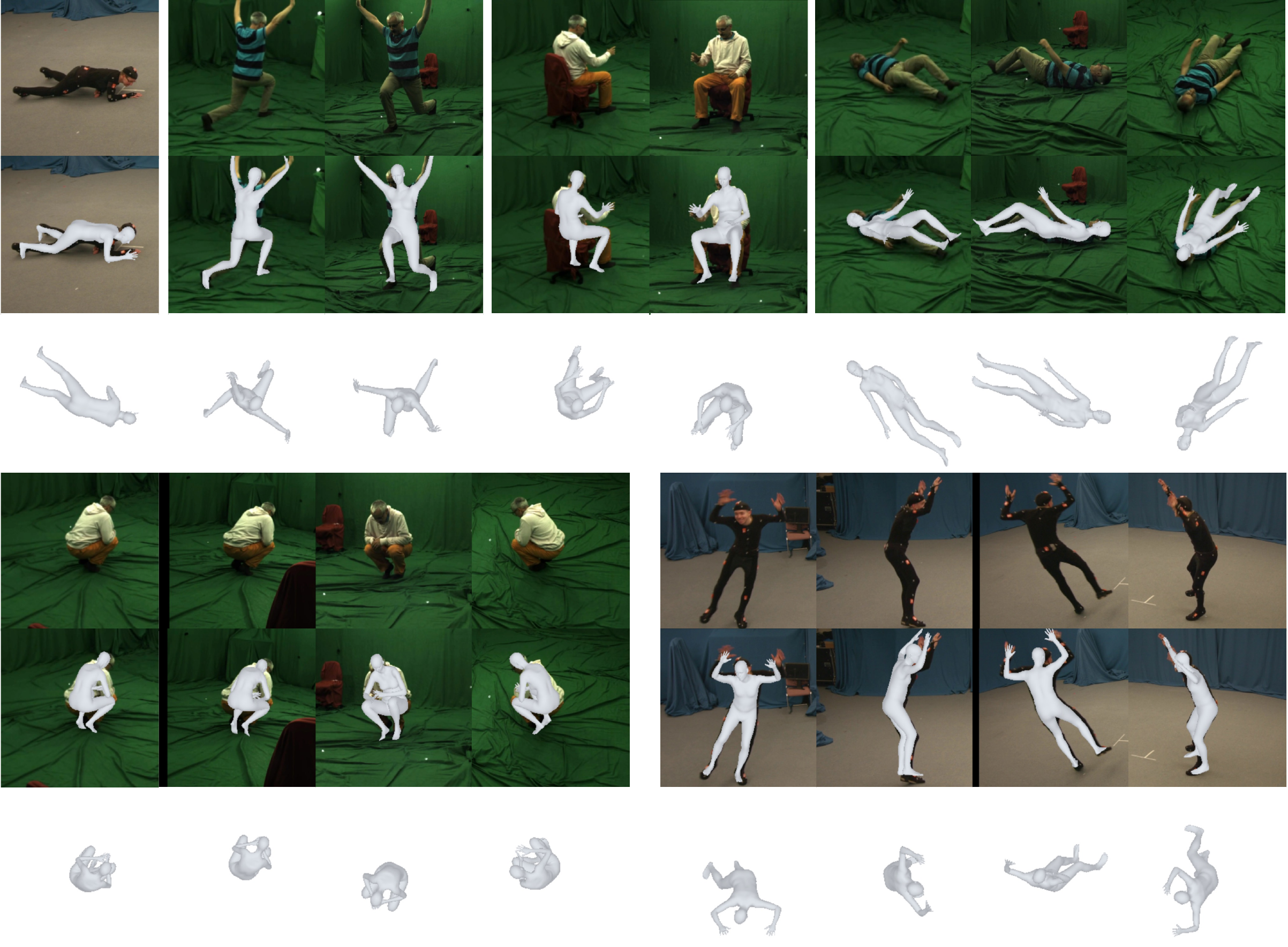}
    \caption{{\bf Visualization results of human mesh recovery from unseen camera views.} We show the results from different number (1-4) of totally unseen views on MPI-INF-3DHPand TotalCapture datasets. For each sample, we show the input image, the reconstruction overlay and the top view.}
    \label{fig:supp_unseen}
\end{figure*}
Our method can recover human mesh from arbitrary multi-view images, including images from totally unseen camera views. For MPI-INF-3DHP and TotalCapture datasets, models are trained on 4 camera views by default as done in the main manuscript. As these two datasets provide more than 4 camera views, we test the model on totally unseen camera views of them, \ie, we randomly select an arbitrary number of unseen camera views and recover human mesh from them. The results are shown in Figure~\ref{fig:supp_unseen}. We can observe that our method performs well even on totally unseen camera views.
\begin{figure*}
    \centering
    \includegraphics[width=\linewidth]{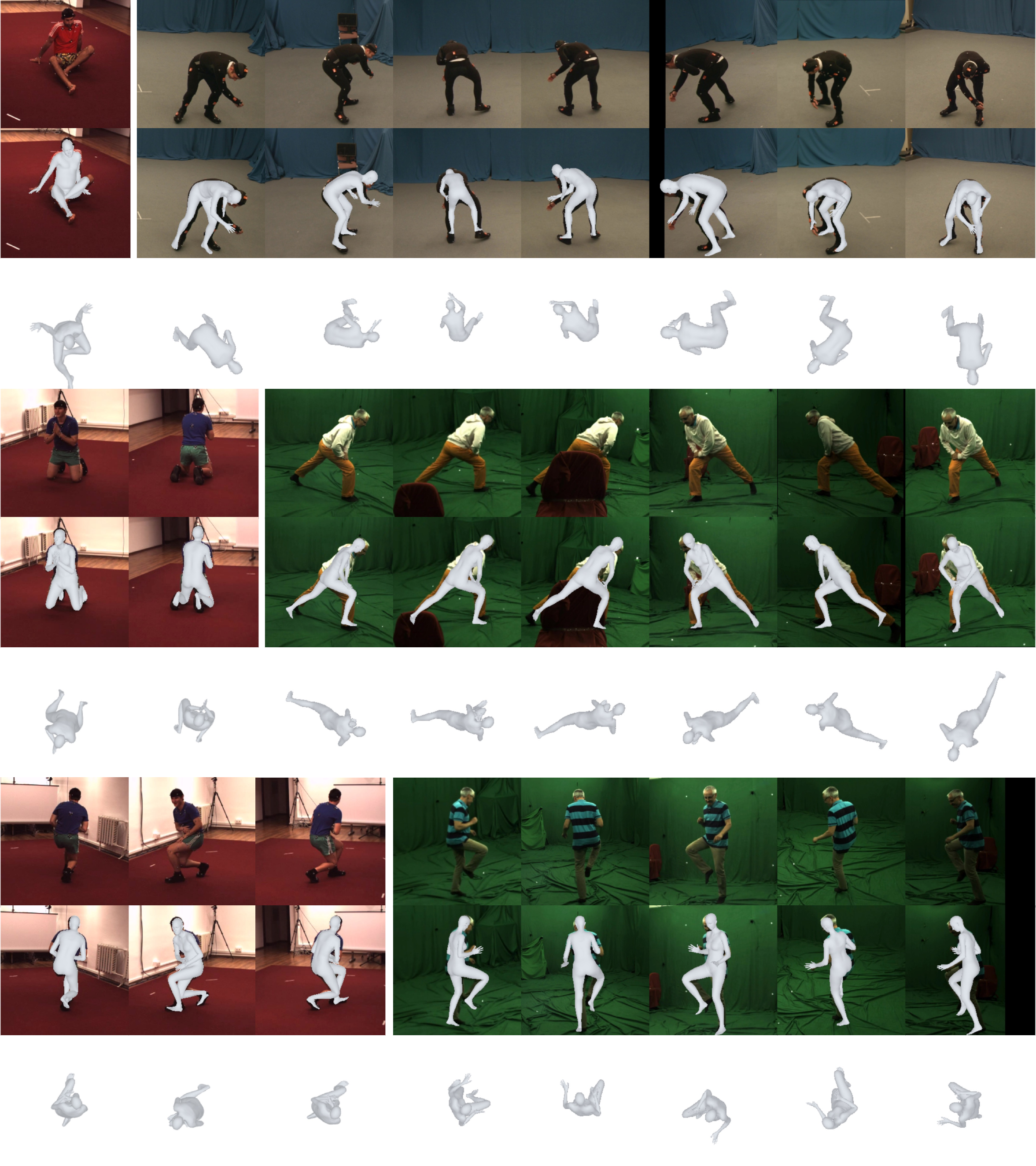}
    \caption{{\bf Visualization results of human mesh recovery from arbitrary multi-view images.} Results from 1-3 and 5-7 camera views on different datasets are shown. For each sample, we show the input image, the reconstruction overlay and the top view.}
    \label{fig:supp_8_1}
\end{figure*}

\begin{figure*}
    \centering
    \includegraphics[width=\linewidth]{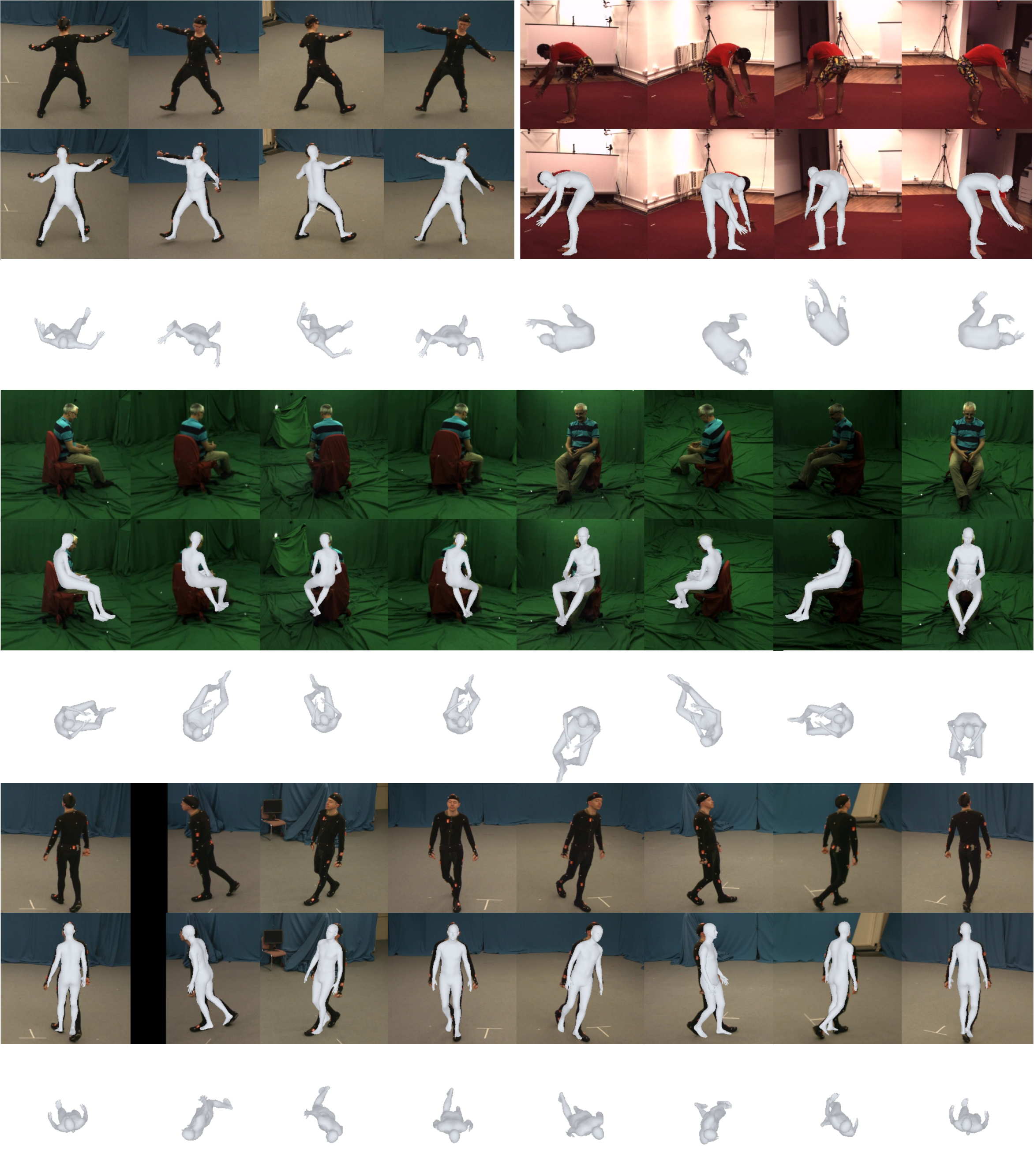}
    \caption{{\bf Visualization results of human mesh recovery from arbitrary multi-view images.} Results from 4 and 8 camera views on different datasets are shown. For each sample, we show the input image, the reconstruction overlay and the top view.}
    \label{fig:supp_8_2}
\end{figure*}
\section{Arbitrary Multi-view Images} \label{sec:supp_2}
In this section, we show visualization results of human mesh recovery from arbitrary multi-view images. As described in the main manuscript, the term ``arbitrary'' involves two characteristics: the arbitrary angle and arbitrary number of views.  Thus, for each of the three datasets (Human3.6M, MPI-INF-3DHP and TotalCapture), we randomly select an arbitrary number of all available (seen and unseen) camera views and recovery human mesh from them. The results are show in Figure~\ref{fig:supp_8_1} and Figure~\ref{fig:supp_8_2}. The results demonstrate the flexibility of our method.

% \begin{figure*}
%     \centering
%     \includegraphics[width=\linewidth]{figures/supp_unseen.pdf}
%     \caption{Caption}
%     \label{fig:supp_unseen}
% \end{figure*}
% \clearpage
% {
%   \small
%   \bibliographystyle{ieeenat_fullname}
%   \bibliography{main}
% }
% \end{document}

% ---- Bibliography ----
%
% BibTeX users should specify bibliography style 'splncs04'.
% References will then be sorted and formatted in the correct style.
%
% \bibliographystyle{splncs04}
% \bibliography{main}
\end{document}